\documentclass{article} 
\usepackage{iclr2016_conference,times}

\iclrfinalcopy

\usepackage{hyperref}
\usepackage{url}
\usepackage{algorithmic,algorithm}
\usepackage{color}
\usepackage{multirow}
\usepackage{amsmath}
\usepackage{amssymb}
\usepackage{graphicx}
\usepackage{subcaption}
\usepackage{here}

\renewcommand{\algorithmiccomment}[1]{\bgroup//~#1\egroup}

\newlength\myindent
\newenvironment{tight_enumerate}{
\begin{enumerate}
  \setlength{\itemsep}{0pt}
  \setlength{\parskip}{0pt}
  \setlength{\topsep}{0pt}
  \setlength{\partopsep}{0pt}
}{\end{enumerate}}

\newenvironment{tight_itemize}{
\begin{itemize}
  \setlength{\itemsep}{0pt}
  \setlength{\parskip}{0pt}
  \setlength{\topsep}{0pt}
  \setlength{\partopsep}{0pt}
}{\end{itemize}}

\title{Distributional Smoothing \\with Virtual Adversarial Training}
\author{Takeru Miyato$^1$, Shin-ichi Maeda$^1$, Masanori Koyama$^1$, Ken Nakae$^1$ \& Shin Ishii$^2$\\
Graduate School of Informatics\\
Kyoto University\\
Yoshidahonmachi 36-1, Sakyo, Kyoto, Japan\\
$^1$\texttt{\{miyato-t,ichi,koyama-m,nakae-k\}@sys.i.kyoto-u.ac.jp}\\
$^2$\texttt{ishii@i.kyoto-u.ac.jp}}

\begin{document}

\maketitle

\begin{abstract}
We propose local distributional smoothness (LDS), a new notion of smoothness for statistical model that can be used as a regularization term to promote the smoothness of the model distribution. We named the LDS based regularization as virtual adversarial training (VAT). 
The LDS of a model at an input datapoint is defined as the KL-divergence based robustness of the model distribution against local perturbation around the datapoint. VAT resembles adversarial training, but distinguishes itself in that it determines the adversarial direction from the model distribution alone without using the label information, making it applicable to semi-supervised learning. The computational cost for VAT is relatively low. For neural network, the approximated gradient of the LDS can be computed with no more than three pairs of forward and back propagations.  When we applied our technique to supervised and semi-supervised learning for the MNIST dataset, it outperformed all the training methods other than the current state of the art method, which is based on a highly advanced generative model. We also applied  our method to SVHN and NORB, and confirmed our method's superior performance over the current state of the art semi-supervised method applied to these datasets. 
\end{abstract}

\section{Introduction}

Overfitting is a \textcolor{black}{serious} problem in supervised training of classification and regression functions. When the number of training samples is finite, training error computed from empirical distribution of the training samples is bound to be different from the test error, which is the expectation of the log-likelihood with respect to the true underlying probability measure ~\cite[]{akaike1998information,watanabe2009algebraic}. 

One popular countermeasure against overfitting is addition of \textcolor{black}{a} regularization term to the objective function. 
\textcolor{black}{The introduction of the regularization term makes the optimal parameter for the objective function to be less dependent on the likelihood term.} In Bayesian framework, \textcolor{black}{the} regularization term corresponds to the logarithm of the prior distribution of the parameters, which defines the preference of the model distribution. Of course, there is no universally good model distribution, and the good model distribution should be chosen dependent on the problem we tackle.
Still yet, our experiences often dictate that the outputs of good models should be smooth with respect to inputs. For example, images and time series occurring in nature tend to be smooth with respect to the space and time~\cite[]{wahba1990spline}.
We would therefore \textcolor{black}{invent a novel regularization term called local distributional smoothness (LDS), } 
which rewards the smoothness of the \textcolor{black}{model} distribution with respect to the input around every \textcolor{black}{input} datapoint,

We define LDS as the negative of the sensitivity of the model distribution $p(y|x,\theta)$ with respect to the perturbation of $x$, measured in the sense of KL divergence. The objective function based on this regularization term is therefore the log likelihood of the dataset augmented with the sum of LDS computed at every \textcolor{black}{input datapoint} in the dataset. 
Because LDS is measuring the local smoothness of the model distribution itself, the regularization term is parametrization invariant. More precisely, our regularized objective function $T$ satisfies the natural property that, if the $\theta^* =  \arg\max_\theta T(\theta) $, then $f(\theta^*) =  \arg\max_\theta T(f(\theta))$ for any diffeomorphism $f$. Therefore, regardless of the parametrization, the optimal model distribution trained with LDS regularization is unique.  This is a property that is not enjoyed by the popular $L_q$ regularization \cite[]{friedman2001elements}. Also, for sophisticated models like deep neural network~\cite[]{bengio2009learning,lecun2015deep},  it is not a simple task to assess the effect of \textcolor{black}{the} $L_q$ regularization term on the topology of the model distribution.  

Our work is closely related to adversarial training~\cite[]{goodfellow2014explaining}. 
At each step of the training, Goodfellow et al. identified for each pair of the observed input $x$ and its label $y$ the direction of the input perturbation to which the classifier's label assignment of $x$ is most sensitive. Goodfellow et al. then penalized the model's sensitivity with respect to the perturbation in the adversarial direction. On the other hand, our LDS defined without label information. Using the language of adversarial training, LDS at each point is therefore measuring the robustness of the model against the perturbation in local and `virtual' adversarial direction. We therefore refer to our regularization method as virtual adversarial training (VAT). Because LDS does not require the label information, VAT is also applicable to semi-supervised learning. This is not the case for adversarial training. 

\textcolor{black}{Furthermore, with \textcolor{black}{the} second order \textcolor{black}{Taylor} expansion of the LDS and an application of power method, we made it possible to approximate the gradient of the LDS efficiently.  The approximated gradient of the LDS can be computed with no more than three pairs of forward and back propagations.}
  
\textcolor{black}{We summarize the advantages of our method below:}
\begin{tight_itemize}
\item \textcolor{black}{Applicability to both supervised and semi-supervised training.}
\item \textcolor{black}{At most two hyperparameters.}
\item \textcolor{black}{Parametrization invariant formulation. The performance of the method is invariant under reparatrization of the model.}
\item \textcolor{black}{Low computational cost.  For Neural network in particular, the approximated gradient of the LDS can be computed with no more than three pairs of forward and back propagations.} 
\end{tight_itemize}
\textcolor{black}{
When we applied the VAT to the supervised and semi-supervised learning of the permutation invariant task for the MNIST dataset, our method outperformed all the contemporary methods other than the state of the art method~\cite[]{rasmus2015semi} that uses a highly advanced generative model based method.  We also applied  our method for semi-supervised learning of permutation invariant task for the SVHN and NORB dataset, and confirmed our method's superior performance over the current state of the art semi-supervised method applied to these datasets. }

\section{Methods}
\subsection{Formalization of Local Distributional Smoothness}
We begin with the formal definition of the local distributional smoothness. 
Let us fix $\theta$ for now, suppose the input space $ \Re ^I $, the output space $Q$, and a training samples
$$D = \{ (x^{(n)}, y^{(n)}) |  x^{(n)} \in \Re ^I , y^{(n)} \in Q, n=1,\cdots, N \},$$ 
and consider the problem of using $D$ to  
train the model distribution $p(y|x,\theta)$ parametrized by $\theta$.
Let ${\rm KL}[p||q]$ denote the KL divergence 
between the distributions $p$ and $q$. Also, with the hyperparameter $\epsilon > 0$, we define
\begin{align}
 \Delta_{\rm KL}(r,x^{(n)},\theta) &\equiv  {\rm KL}[p(y|x^{(n)},\theta)\|p(y|x^{(n)}+ r,\theta)]\label{eq:divergence} \\
r_{\rm v\text{-}adv}^{(n)} &\equiv \arg \mathop {\rm max}\limits_r  \{ \Delta_{\rm KL}(r,x^{(n)},\theta); ~ \|r\|_2 \leq \epsilon \}. \label{eq:rvadv}
\end{align}
From now on, we refer to 
$r_{\rm v\text{-}adv}^{(n)}$ as the virtual adversarial perturbation. 
We define the local distributional smoothing (LDS) of the model distribution at $x^{(n)}$ by  
\begin{eqnarray}
{\rm LDS }(x^{(n)},\theta) &\equiv& -\Delta_{\rm KL}(r_{\rm v\text{-}adv}^{(n)},x^{(n)},\theta). \label{eq:ds} 
 \label{eq:LDSdef}
\end{eqnarray}
Note $r_{\rm v\text{-}adv}^{(n)}$ is the direction to which the model distribution 
$p(y|x^{(n)},\theta)$ is most sensitive in the sense of KL divergence. In a way, this is a 
KL divergence analogue of the gradient $\nabla_x$ of the model distribution with respect to the input, 
and perturbation of $x$ in this direction wrecks the local smoothness of $p(y|x^{(n)},\theta)$ at $x^{(n)}$ in a
most dire way.  The smaller the value of $ \Delta_{\rm KL}(r_{\rm v\text{-}adv}^{(n)},x^{(n)},\theta) $ at $x^{(n)}$, the smoother the $p(y|x^{(n)},\theta)$ at $x$. Our goal is to improve the smoothness of the model in the neighborhood of all the observed inputs. Formulating this goal based on the LDS, we obtain the following objective function, 
\begin{eqnarray}
\frac{1}{N}\sum_{n=1}^N \log p(y^{(n)}|x^{(n)},\theta) 
+ \lambda \frac{1}{N}\sum_{n=1}^N{\rm LDS} (x^{(n)},\theta). \label{eq:LDSMAPobj}
\end{eqnarray}
We call the training based on \eqref{eq:LDSMAPobj} the virtual adversarial training (VAT). By the construction, VAT is parametrized by the hyperparameters $\lambda>0$ and 
$\epsilon>0$. 
If we define  $ r_{\rm adv}^{(n)} \equiv \arg \mathop {\rm min}\limits_r  \{p(y^{(n)}|x^{(n)}+ r,\theta),\|r\|_p \leq \epsilon \} $ and replace $-\Delta_{\rm KL}(r_{\rm v\text{-}adv}^{(n)},x^{(n)},\theta)$ in \eqref{eq:ds} with  $\log p(y^{(n)}|x^{(n)}+ r_{\rm adv}^{(n)},\theta)$, 
we obtain the objective function of the adversarial training~\cite[]{goodfellow2014explaining}. Perturbation of $x^{(n)}$ in the direction of $r_{\rm adv}^{(n)}$ can most severely damage the probability that the model correctly assigns the label $y^{(n)}$ to $x^{(n)}$.  As opposed to $r_{\rm adv}^{(n)}$, the definition of $r_{\rm v\text{-}adv}^{(n)}$ on $x^{(n)}$ does not require the correct label $y^{(n)}$.  This property allows us to apply the VAT to semi-supervised learning. 

LDS is a definition meant to be applied to any model with distribution that are
smooth with respect to $x$.  For instance, for a 
linear regression model $p(y|x,\theta)=\mathcal{N}(\theta^{\rm T}x, \sigma^2)$
the LDS becomes
\begin{align}
{\rm LDS}(x,\theta)  &= -\frac{1}{2\sigma^2} \epsilon ^2 \|\theta\|_2^2 , \nonumber
\end{align}
and this is the same as the form of $L_2$ regularization. This does not, however, mean that 
$L_2$ regularization and LDS is equivalent for linear models. 
It is not difficult to see that, when we reparametrize the model as $p(y|x,\theta)=\mathcal{N}({\theta^{3}}^{\rm T} x, \sigma^2),$ we obtain ${\rm LDS}(x,\theta^3) \propto -\epsilon ^2 \|\theta^3\|_2^2$, not $-\epsilon ^2 \|\theta\|_2^2$. 
For a logistic regression model 
$p(y=1|x,\theta)= \sigma(\theta^{\rm T}x) = (1+\exp(-\theta^{\rm T}x))^{-1}$, 
we obtain
\begin{align}
 {\rm LDS}(x,\theta) &\cong -\frac{1}{2} \sigma(\theta^{\rm T}x)\bigl(1-\sigma(\theta^{\rm T}x)\bigr)\epsilon^2\|\theta\|^2_2 \nonumber
\end{align}
with the second-order Taylor approximation with respect to $\theta^T r$.

\subsection{Efficient evaluation of LDS and its derivative with respect to $\theta$}
Once $r_{\rm v\text{-}adv}^{(n)}$ is computed, the evaluation of the LDS is simply the computation of the KL divergence between the model distributions $p(y |x^{(n)},\theta)$ and $ p(y|x^{(n)} + r_{\rm v\text{-}adv}^{(n)},\theta)$. When $p(y |x^{(n)},\theta)$ can be approximated with well known exponential family, this computation 
is straightforward. For example, one can use Gaussian approximation for many cases of NNs. 
In what follows, we discuss the efficient computation of $r_{\rm v\text{-}adv}^{(n)}$, for which there is no evident approximation. 

\subsubsection{Evaluation of $r_{\rm v\text{-}adv}$} 
We assume that $p(y |x,\theta)$ is differentiable with respect to $\theta$ and $x$ almost everywhere.
Because $\Delta_{\rm KL}(r,x,\theta)$ takes minimum value at $r = 0$, 
the differentiability assumption dictates that 
its first derivative $\nabla_r \Delta_{\rm KL}(r,x,\theta)|_{r=0}$ is zero.  Therefore we can take the second-order Taylor approximation as 
\begin{eqnarray}
\Delta_{\rm KL}(r,x,\theta) \cong  \frac{1}{2}r^{T}H(x,\theta) r, \label{eq:second_taylor}
\end{eqnarray}
where $ H(x,\theta)$ is a Hessian matrix given by $ H(x,\theta) \equiv \nabla\nabla_{r} \Delta_{\rm KL}(r,x,\theta)|_{r=0}$. 
Under this approximation $r_{\rm v\text{-}adv}$ emerges as the first dominant eigenvector of $H(x,\theta)$,
$u(x,\theta)$, of magnitude $\epsilon$, 
\begin{eqnarray}
r_{\rm v\text{-}adv}(x,\theta) &\cong& \arg \mathop {\rm max}\limits_r  \{r^{T} H(x,\theta) r; ~\|r\|_2 \leq \epsilon\}  \nonumber \\
&=& \epsilon \overline{u(x,\theta)}, 
\end{eqnarray} 
where $\bar{\cdot}$ denotes an operator acting on arbitrary non-zero vector $v$
that returns a unit vector in the direction of $v$ as $\bar{v}$.
Hereafter, we denote
$H(x,\theta)$ and $u(x,\theta)$ as 
$H$ and $u$, respectively.


The eigenvectors of the Hessian $H(x, \theta)$ require $O(I^3)$ computational time, which becomes unfeasibly large for high dimensional input space. We therefore resort to power iteration method~\cite[]{golub2000eigenvalue} and finite difference method to approximate $r_{\rm v\text{-}adv}$.  
Let $d$ be a randomly sampled unit vector. As long as 
$d$ is not perpendicular to the dominant eigenvector $u$,  
the iterative calculation of
\begin{eqnarray}
 d \leftarrow \overline{H d} \label{eq:power_method}
 \end{eqnarray}
will make the $d$ converge to $u$.  
We need to do this without the direct computation of $H$, however. 
$H d$ can be approximated by finite difference \footnote{\textcolor{black}{For many models including neural networks, $Hd$ can be computed exactly \cite[]{pearlmutter1994fast}. We forgo this computation in this very paper, however, because the implementation of his procedure is not straightforward in some of the standard deeplearning frameworks (e.g. Caffe~\cite[]{jia2014caffe}, Chainer~\cite[]{tokuichainer}).}}
\begin{eqnarray}
H d &\cong& \frac{\nabla_{r} \Delta_{\rm KL}(r,x,\theta) |_{r=\xi d} - \nabla_{r} \Delta_{\rm KL}(r,x,\theta)|_{r=0}}{\xi} \nonumber \\
&=& \frac{\nabla_{r} \Delta_{\rm KL}(r,x,\theta)|_{r=\xi d}}{\xi}, \label{eq:dif_method}
\end{eqnarray}
with $\xi \neq 0$. In the computation above, we used the fact $\nabla_r \Delta_{\rm KL}(r,x,\theta)|_{r=0} = 0$ again.  In summary, 
we can approximate $r_{\rm v\text{-}adv}$ with the repeated application of the following update: 
\begin{eqnarray}
d \leftarrow \overline{\nabla_{r} \Delta_{\rm KL}(r,x,\theta)|_{r=\xi d}} .
\end{eqnarray}
The approximation improves monotonically 
with the iteration times of the power method, $I_p$.  
Most notably, the value $\nabla_{r} \Delta_{KL}(r+\xi d,x,\theta)|_{r=0}$ can be computed easily by the back propagation method in the case of neural networks. We denote the approximated $r_{\rm v\text{-}adv}^{(n)}$ as 
$\tilde r_{\rm v\text{-}adv}^{(n)} = {\rm GenVAP}(\theta,x^{(n)},\epsilon,{\rm I}_p,\xi)$ (See Algorithm~\ref{alg:gen_vap}), and denote the LDS computed with 
$\tilde r_{\rm v\text{-}adv}^{(n)}$ as 
\begin{equation}
\widetilde{\rm LDS}(x^{(n)},\theta) \equiv-\Delta_{\rm KL}(\tilde{r}_{\rm v\text{-}adv}^{(n)},x^{(n)},\theta).
\end{equation}

\begin{algorithm}
\caption{Generation of $\tilde r_{\rm v\text{-}adv}^{(n)}$}\label{alg:gen_vap}
\begin{algorithmic}[]
\STATE\textbf{Function} GenVAP($\theta$,$x^{(n)}$,$\epsilon$,$I_p$,$\xi$)
\begin{tight_enumerate}
\item Initialize $d  \in R^{I}$  by a random unit vector.
\item \textbf{Repeat For} i in $1 \dots I_p$ (Perform $ I_p $-times power method)
\begin{tight_enumerate}
\item[] $d \leftarrow \overline{\nabla_{r} \Delta_{\rm KL}(r,x^{(n)},\theta)|_{r=\xi d}}$
\end{tight_enumerate}
\item \textbf{Return} $\epsilon d$
\end{tight_enumerate}
\end{algorithmic}
\end{algorithm}

\subsubsection{Evaluation of derivative of approximated LDS w.r.t $\theta$}

Let $\hat \theta$ be the current value of $\theta$ in the algorithm.
It remains to evaluate the derivative of $\widetilde{\rm LDS}(x^{(n)},\theta)$ with respect to $\theta$ at 
$\theta = \hat \theta$, or
\begin{align}
\begin{split}
\left. {\frac{\partial }{\partial \theta } \widetilde{\rm LDS}(x^{(n)},\theta)} \right|_{\theta = \hat \theta}
= -\left. {\frac{\partial }{\partial \theta }{\rm KL}\big[p\big(y|x^{(n)},\theta\big)\|p\big(y|x^{(n)}+\tilde{r}_{\rm v\text{-}adv}^{(n)},\theta\big) \big] } \right|_{\theta = \hat \theta}.
\end{split} \label{eq:LDS_grad}
\end{align}
\textcolor{black}{
By \textcolor{black}{the} definition, $\tilde r_{\rm  v\text{-}adv}$ depends on $\theta$. Our numerical experiments, however, indicates that $\nabla _{\theta}r_{\rm v\text{-}adv}$ is quite volatile with respect to $\theta$, and we could not make effective regularization 
when we used the numerical evaluation of $\nabla_{\theta} \tilde{r}_{\rm v\text{-}adv}$
in $\nabla_{\theta} \widetilde{\rm LDS}(x^{(n)},\theta)$. 
We have therefore followed the work of \cite{goodfellow2014explaining} and ignored 
the derivative of $\tilde{r}_{\rm v\text{-}adv}$ with 
respect to $\theta$. This modification in fact achieved better generalization performance and higher $\widetilde{\rm LDS}(x^{(n)},\theta)$. 
We also replaced the first $\theta$ in the KL term of  \eqref{eq:LDS_grad} with $\hat \theta$ and computed 
 \begin{eqnarray}
 -\left. {\frac{\partial }{\partial \theta }{\rm KL}\big[p\big(y|x^{(n)},\hat{\theta}\big)\|p\big(y|x^{(n)}+\tilde{r}_{\rm v\text{-}adv}^{(n)},\theta\big) \big] } \right|_{\theta = \hat \theta}.
\label{eq:derivative of approximate LDS}
\end{eqnarray}
The stochastic gradient descent based on ~\eqref{eq:LDSMAPobj} with \eqref{eq:derivative of approximate LDS} was able to achieve even better generalization performance. 
From now on, we refer to the training of the 
regularized likelihood  \eqref{eq:LDSMAPobj}  based on ~\eqref{eq:derivative of approximate LDS} 
as virtual adversarial training (VAT).}


\subsection{\textcolor{black}{Computational cost of computing the gradient of LDS}}

{\color{black}
We would like to also comment on the computational cost required for \eqref{eq:derivative of approximate LDS}. We will restrict our discussion to the case with $I_p =1$ in \eqref{eq:power_method} , because one iteration of power method was sufficient for computing accurate $Hd$ and increasing $I_p$ did not have much effect in all our experiments. 

With the efficient computation of LDS we presented above, we only need to compute $\nabla_{r} \Delta_{\rm KL}(r+\xi d,x^{(n)},\theta)$ in order to compute $r_{\rm v\text{-}adv}^{(n)}$. Once $r_{\rm v\text{-}adv}^{(n)}$ is decided, we compute the gradient of the LDS with respect to the parameter with \eqref{eq:derivative of approximate LDS}.  
For neural network, the steps that we listed above require only two forward propagations and two back propagations. 
In semi-supervised training, we would also need to evaluate the probability distribution $p(y|x^{(n)},\theta)$ in \eqref{eq:divergence} for unlabeled samples. As long as we use the same dataset to compute the likelihood term and the LDS term, this requires only one additional forward propagation.  Overall, especially for neural network, we need no more than three pairs of forward propagation and back propagation to compute the derivative approximated LDS with \eqref{eq:derivative of approximate LDS}.}

\section{Experiments\label{sec:experiments}}
All the computations were conducted with Theano~\cite[]{bergstra+al:2010-scipy,Bastien-Theano-2012}. Reproducing code is uploaded on \url{https://github.com/takerum/vat}. Throughout the experiments on our proposed method, we used a fixed value of $\lambda = 1$, and we also used a fixed value of $I_p = 1$ except for the experiments of synthetic datasets.

\subsection{Supervised learning for the binary classification of synthetic dataset}
\begin{figure}[h]
\begin{center}
\begin{subfigure}[b]{0.22\textwidth}
\includegraphics[width=1.0\textwidth]{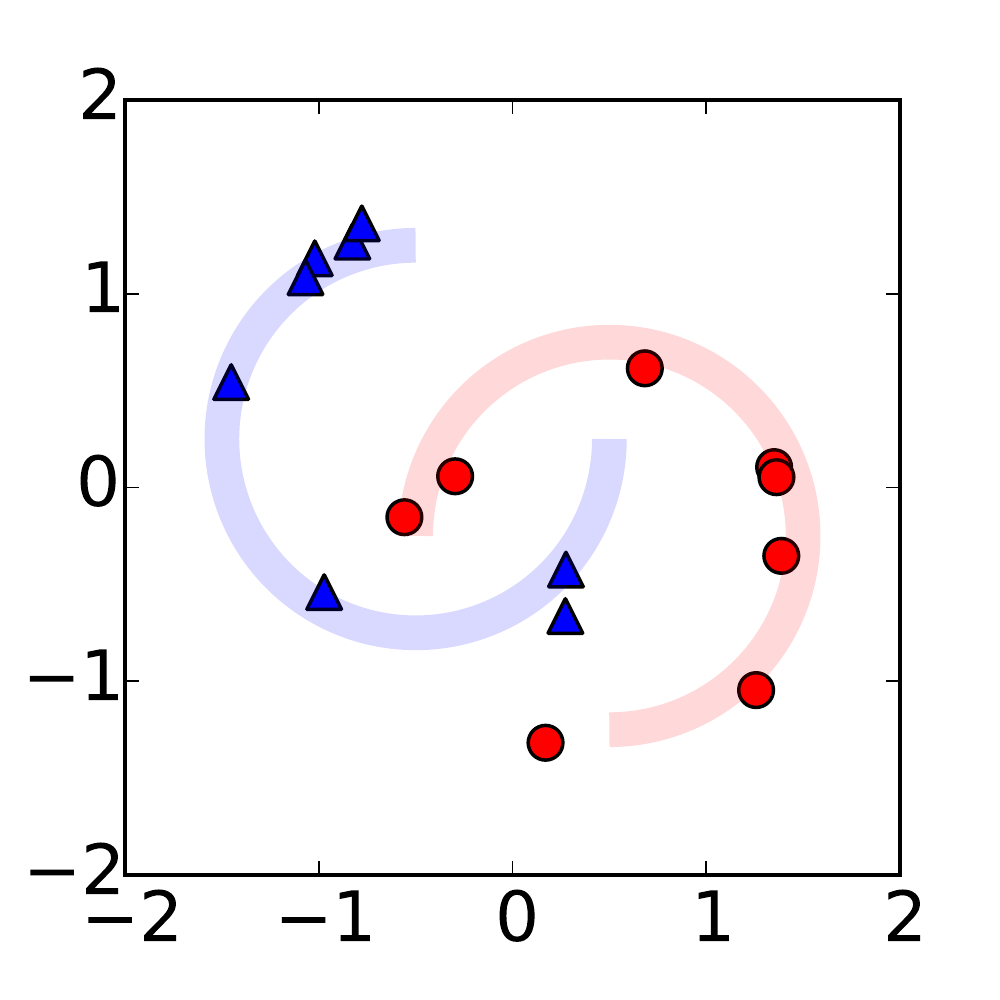}
\caption{Moons dataset}
\end{subfigure}%
\begin{subfigure}[b]{0.22\textwidth}
\includegraphics[width=1.0\textwidth]{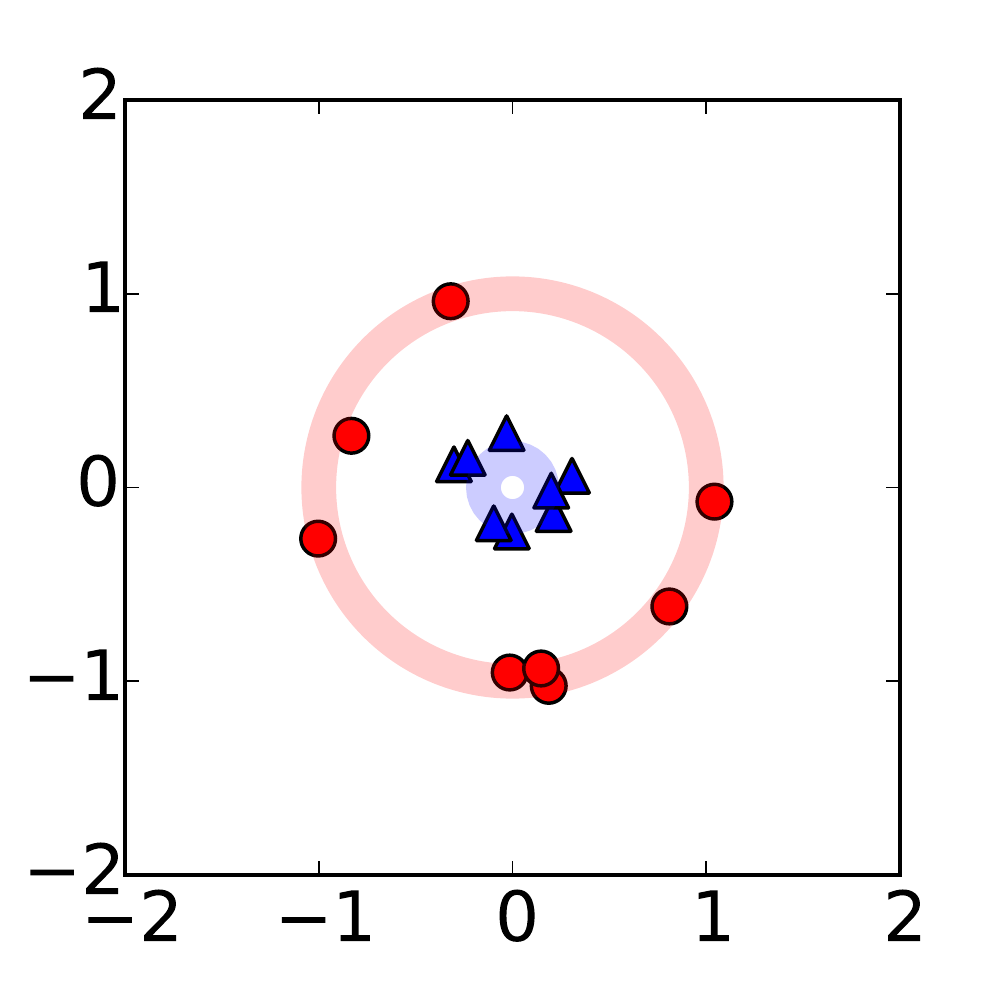}
\caption{Circles dataset}
\end{subfigure}%
\caption{\label{fig:synthetic_datasets}Visualization of the synthetic datasets. Each panel depicts the total of 16 training data points. 
Red circles stand for the samples with label $1$ and blue triangles stand for the samples with label $0$.
Samples with each label are prepared uniformly from the light-colored trajectory.}
\end{center}
\end{figure}
We created two synthetic datasets by generating multiple points uniformly over two trajectories on $\Re^{2}$ as shown in Figure \ref{fig:synthetic_datasets} and linearly embedding them into 100 dimensional vector space. 
The datasets are called (a) `Moons' dataset and (b) `Circles' dataset based on the shapes of the two trajectories, respectively.

Each dataset consists of 16 training samples and 1000 test samples. Because the number of the samples  is very small relative to the input dimension,
maximum likelhood estimation (MLE) is vulnerable to overfitting problem on these datasets.
The set of hyperparameters in each regularization method and the other detailed experimental settings are described in Appendix~\ref{apd:syn_exp_set}.
We repeated the experiments 50 times with different samples of training and test sets, and reported the average of the 50 test performances.

Our classifier was a neural network (NN) with one hidden layer consisting of 100 hidden units. We used ReLU \cite[]{jarrett2009best,nair2010rectified,glorot2011deep} activation function for hidden units, and used softmax activation function for all the output units. 
The regularization methods we compared against the VAT on this dataset include $L_2$ regularization ($L_2$-reg), dropout~\cite[]{srivastava2014dropout}, adversarial training(Adv), and random perturbation training (RP). The random perturbation training is a modified version of the VAT in which we replaced $r_{\rm v\text{-adv}}$ with an $\epsilon$ sized unit vector uniformly sampled from $I$ dimensional unit sphere.
We compared random perturbation training with VAT in order to highlight the importance of choosing the appropriate direction of the perturbation.  As for the adversarial training, we followed \cite{goodfellow2014explaining} and determined the size of the perturbation $r$ in terms of both $L_{\infty}$ norm and $L_2$ norm.  

\begin{figure}[h]
\begin{center}
\includegraphics[width=1.0\textwidth]{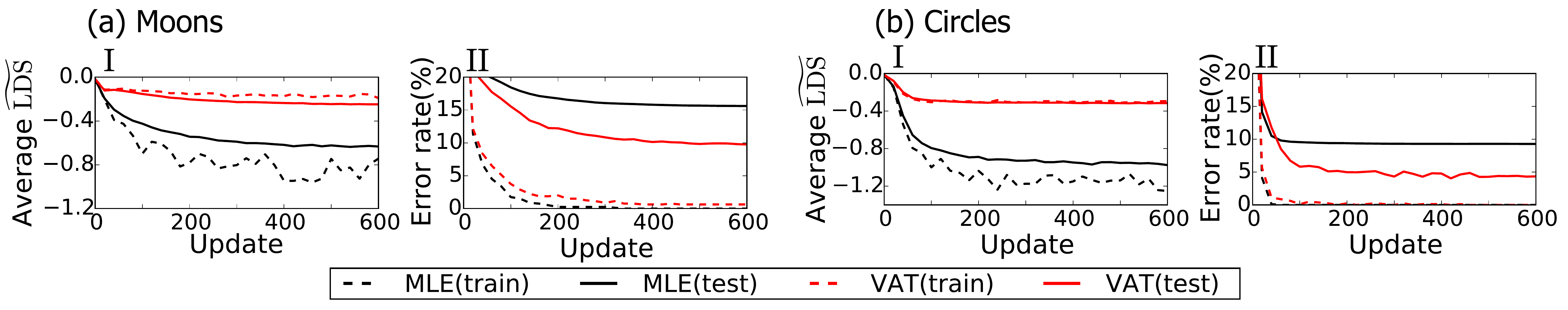}
\caption{\label{fig:training_process_for_synthetic_dataset}
Comparison of transitions of average LDS(I) and error rate(II) between MLE and VAT implemented with $\epsilon = 0.5$ and $I_p=1$. Average $\widetilde{\rm LDS}$ showed in (a.I) and (b.I) were evaluated on the training and test samples with $\epsilon = 0.5$ and $I_p=5$.
}
\end{center}
\end{figure}
Figure \ref{fig:training_process_for_synthetic_dataset} compares the learning process between the VAT and the MLE.    
Panels (a.I) and (b.I) show the transitions of the average $\widetilde{\rm LDS}$,
while panels (a.II) and (b.II) show the transitions of the error rate, on both training and test set.
The average $\widetilde{\rm LDS}$ is nearly zero at the beginning, because the models are initially close to uniform distribution around each inputs. 
The average $\widetilde{\rm LDS}$ then decreases slowly for the VAT, and falls rapidly for the MLE.
Although the training error eventually drops to zero for both methods, the final test error of the VAT is significantly lower than that of the MLE. This difference suggests that a high sustained value of $\widetilde{\rm LDS}$ is beneficial in alleviating the overfitting and in decreasing the test error.

\begin{figure}[h]
\centering
\includegraphics[width=1.0\textwidth]{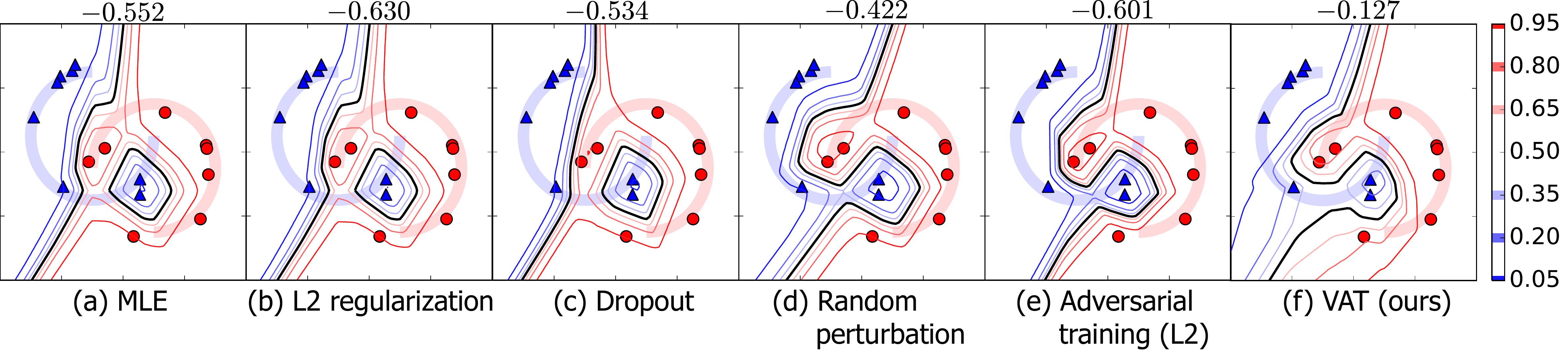}
\caption{\label{fig:pred_dataset_A}Contours of $p(y =1 | x, \theta)$ drawn by NNs (with ReLU activation) trained with various regularization methods for a single dataset of `Moons'.
A black line represents the contour of value $0.5$.
Red circles represent the data points with label 1,
and blue triangles represent the data points with label 0.
The value above each panel correspond to average $\widetilde{\rm LDS}$ value. Average $\widetilde{\rm LDS}$ evaluated on the training set with
$\epsilon = 0.5$ and $I_p=5$ is shown at the top of each panel.}
\end{figure}

\begin{figure}[h]
\centering
\includegraphics[width=1.0\textwidth]{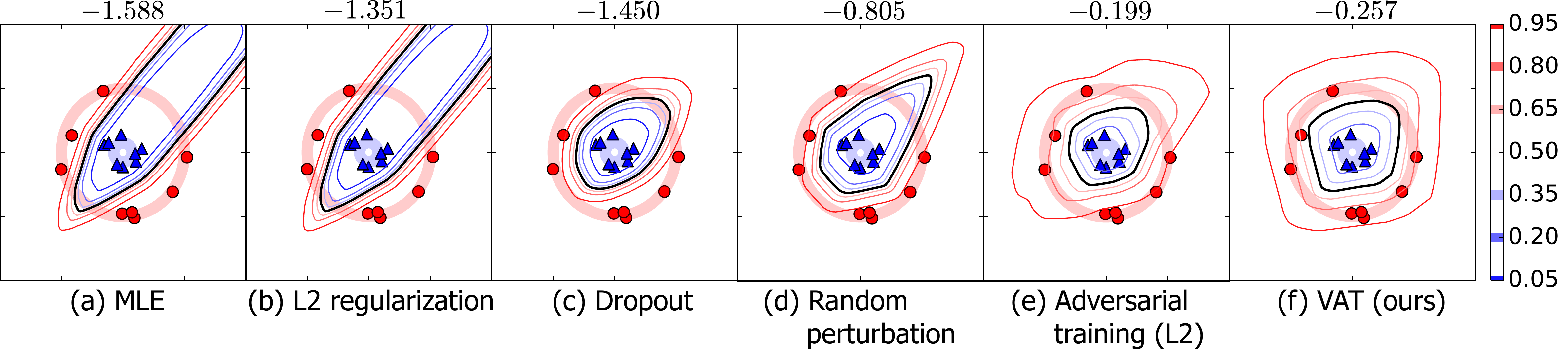}
\caption{\label{fig:pred_dataset_B}Contours of $p(y =1 | x, \theta)$ drawn by NNs trained with various regularization methods for a single dataset of `Circles'. The rest of the details follow the caption of the Figure \ref{fig:pred_dataset_A}.}
\end{figure}

Figures \ref{fig:pred_dataset_A} and \ref{fig:pred_dataset_B} show the contour plot of \textcolor{black}{the} model distributions for the binary classification problems of `Moons' and `Circles' trained with the best set of hyper parameters. The value above each panel correspond to average $\widetilde{\rm LDS}$ value. We see from the figures that NN without regularization (MLE) and 
\textcolor{black}{
NN with $L_2$ regularization are drawing decisively wrong decision boundary. The decision boundary drawn by 
dropout for `Circles' is convincing, but the dropout's decision boundary for `Moons' does not coincide with our intention. The opposite can be said for the random perturbation training. 
Only adversarial training and VAT are consistently yielding the intended decision boundaries for both datasets.  
VAT is drawing appropriate decision boundary by imposing local smoothness regularization around each data point.  } This does not mean, however, that the large value of LDS immediately implies good decision boundary. By its very definition, $\widetilde{\rm LDS}$ tends to disfavor abrupt change of the likelihood around training datapoint. Large value of $\widetilde{\rm LDS}$ therefore forces large relative margin around the decision boundary. One can achieve large value of $\widetilde{\rm LDS}$ with $L_2$ regularization, dropout and random perturbation training with appropriate choice of hyperparameters by smoothing the model distribution globally. This, indeed, comes at the cost of accuracy, however.  
\begin{figure}[h]
\begin{center}
\begin{subfigure}[b]{0.33\textwidth}
\includegraphics[width=1.0\textwidth]{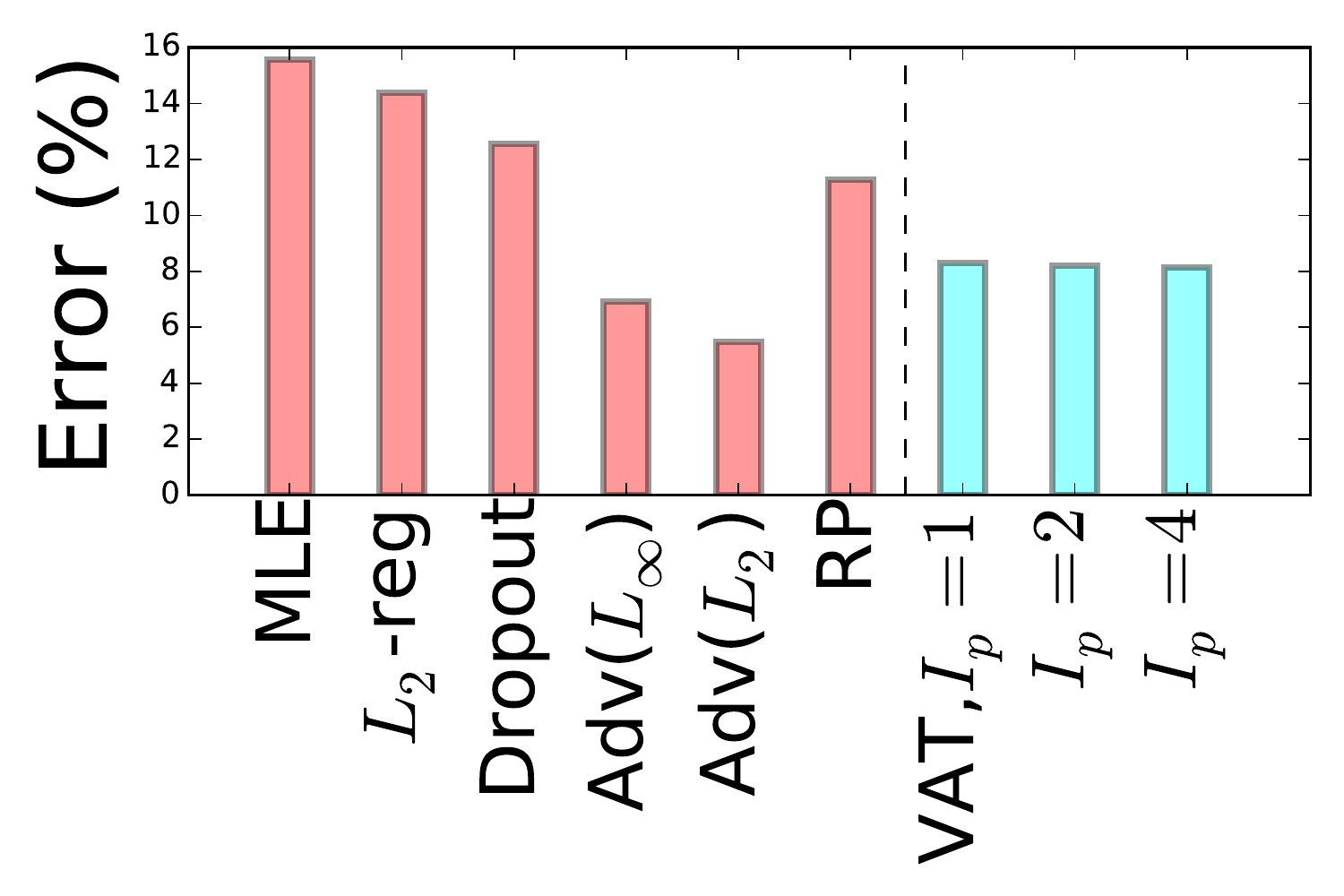}
\caption{Moons}
\end{subfigure}
\begin{subfigure}[b]{0.33\textwidth}
\includegraphics[width=1.0\textwidth]{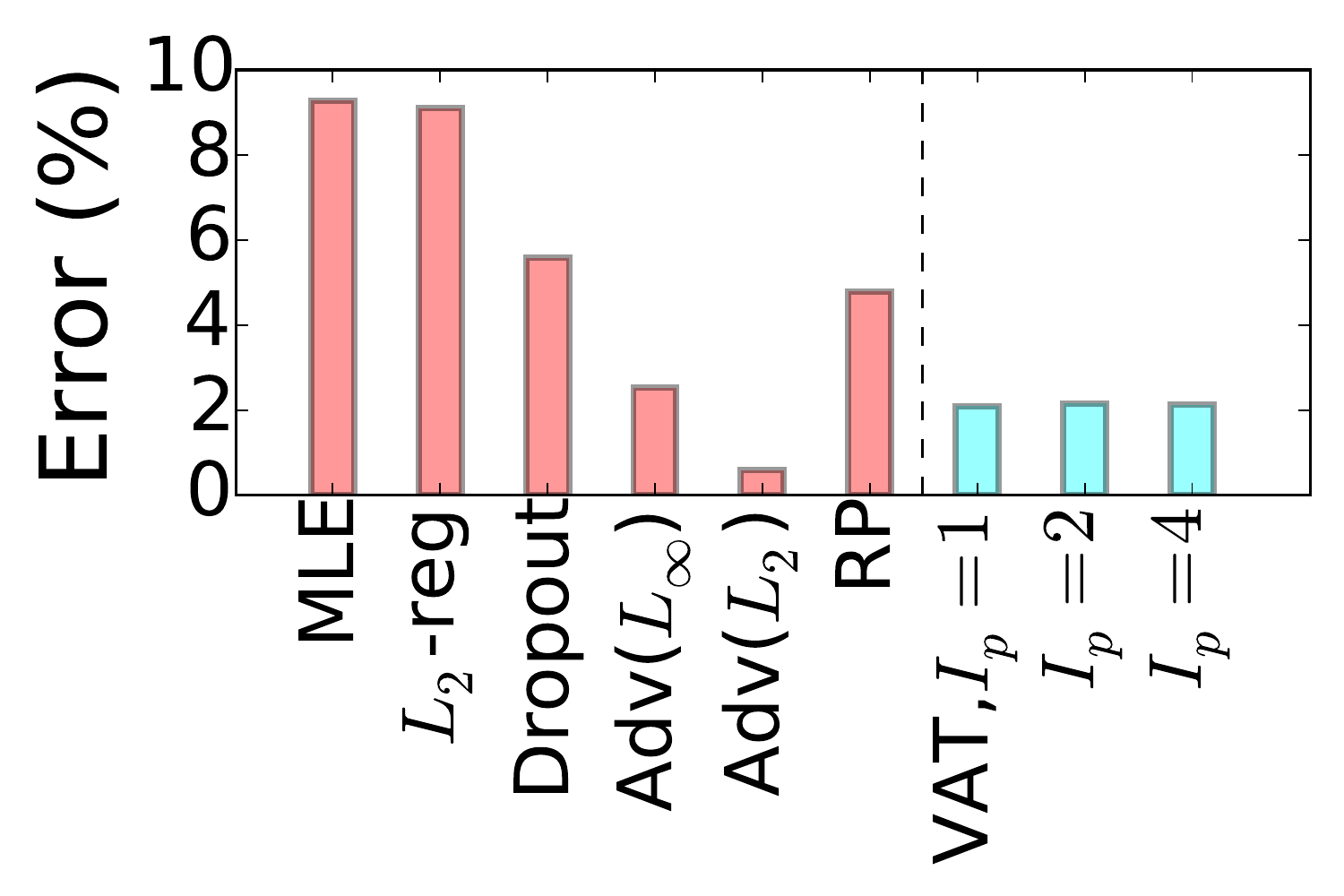}
\caption{Circles}
\end{subfigure}
\caption{\label{fig:comparison_test_errors_for_synthetic_dataset}  Comparison of the test error rate for (a) `Moons' and (b) `Circles'}
\end{center}
\end{figure}
Figure \ref{fig:comparison_test_errors_for_synthetic_dataset} summarizes the average test errors of six regularization methods with the best set of hyperparameters. Adversarial training and VAT achieved much lower test errors than the other regularization methods. Surprisingly, the performance of the VAT did not change much with the value of $I_p$. We see that $I_p=1$ suffices for our dataset.
\subsection{Supervised learning for the classification of the MNIST dataset}\label{subsec:scfMNIST}
Next, we tested the performance of our regularization 
method on the MNIST dataset, which consists of $28 \times 28$ pixel images of handwritten digits and their corresponding labels. The input dimension is therefore $28 \times 28 = 784$ and 
each label is one of the numerals from $0$ to $9$.
We split the original $60{,}000$ training samples into $50{,}000$ training samples and $10{,}000$ validation samples, and used the latter of which to tune the hyperparameters.
We applied our methods to the training of 2 types of NNs with different numbers of layers, 2 and 4. As for the number of hidden units, we used $(1200, 600)$ and $(1200, 600, 300, 150)$ respectively. The ReLU activation function and batch normalization technique~\cite[]{ioffe2015batch} were used for all the NNs. The detailed settings of the experiments are described in Appendix~\ref{apd:mnist_exp_set}.

 
For each regularization method, we used the set of hyperparameters that achieved the best performance on the validation data to train the NN on all training samples.
We applied the trained networks to the test set and recorded their test errors.  We repeated this procedure 10 times with different seeds for the weight initialization, and reported the average test error values.

Table \ref{tab:test_performances} summarizes the test error obtained by our regularization method (VAT) and the other regularization methods.
VAT performed better than all the contemporary methods except Ladder network, which is highly advanced generative model based method. 
\begin{table}[ht]
	\begin{center}
		\caption{\label{tab:test_performances}Test errors of 10-class supervised learning for the permutation invariant MNIST task. Stars * indicate the methods that are dependent on generative models or pre-training. Test errors in the upper panel are the ones reported in the literature, and test errors in the bottom panel are the outputs of our implementation.}
        \scalebox{0.9}{
        \tabcolsep=0.05cm
		\begin{tabular}{lc}
			Method & Test error (\%) \\
          	\hline\hline
            SVM (gaussian kernel) & 1.40 \\
            Gaussian dropout~\cite[]{srivastava2014dropout} & 0.95 \\
            Maxout Networks~\cite[]{goodfellow2013maxout} & 0.94\\
            *MTC~\cite[]{rifai2011manifold} & 0.81 \\
            *DBM~\cite[]{srivastava2014dropout}& 0.79 \\
            Adversarial training~\cite[]{goodfellow2014explaining} & 0.782\\
            *Ladder network~\cite[]{rasmus2015semi} & 0.57$\pm$0.02 \\
            \hline
            Plain NN (MLE) & 1.11 \\
            Random perturbation training & 0.843 \\
            Adversarial training (with $L_{\infty}$ norm constraint) & 0.788\\
            Adversarial training (with $L_2$ norm constraint) & 0.708\\
            VAT (ours) & 0.637$\pm 0.046$ \\
		\end{tabular}
        }
   \end{center}
\end{table}

\subsection{Semi-supervised learning for the classification of the benchmark datasets}

{\color{black}
Recall that our definition of LDS (Eq.\eqref{eq:LDSdef}) at any point $x$ is independent of the label information $y$. This in particular means that we can apply the VAT to semi-supervised learning tasks.  We would like to emphasize that this is a property not enjoyed by adversarial training and dropout.  We applied VAT to semi-supervised learning tasks in the permutation invariant setting for three datasets: MNIST, SVHN, and NORB.
In this section we describe the detail of our setup for the semi-supervised learning of the MNIST dataset, and leave the further details of the experiments in the Appendix~\ref{apd:semi-mnist_exp_set} and~\ref{apd:semi-svhn_norb_exp_set}. 

We experimented with four sizes of labeled training samples $N_l = \{100, 600, 1000, 3000\}$ and observed the effect of $N_l$ on the test error. We used the validation set of fixed size $1000$, and used all the training samples excluding the validation set and the labeled to train the NNs.  That is, when $N_l = 100$, the unlabeled training set had the 
size of $60{,}000 - 100 - 1{,}000 = 58{,}900$. 
For each choice of the hyperparameters, we repeated the experiment 10 times with different set.
As for the architecture of NNs, we used ReLU based NNs with two hidden layers with the number of hidden units $(1200, 1200)$. 
Batch normalization was implemented as well. 

Table \ref{tab:MNIST} summarizes the results for the permutation invariant MNIST task.  
All the methods other than SVM and plain NN (MLE) are semi-supervised learning methods. For the MNIST dataset, VAT outperformed all the contemporary methods other than Ladder network \cite[]{rasmus2015semi}, which is current state of the art.  

Table \ref{tab:SVHN} and Table \ref{tab:NORB} summarizes the the results for SVHN and NORB respectively. Our method strongly outperforms the current state of the art semi-supervised learning method applied to these datasets. 
}

\begin{table}[ht]
	\caption{\label{tab:semi-supervised_test_errors}\textcolor{black}{ Test errors of semi-supervised learning for the permutation invariant task for MNIST, SVHN and NORB. All values are the averages over random partitioning of the dataset. Stars * indicate the methods that are dependent on generative models or pre-training. }}
    \begin{center}
    \scalebox{0.9}{
    \begin{subtable}[h]{1.0\textwidth}
    \centering
    \caption{MNIST \label{tab:MNIST}}
		\begin{tabular}{lrrrrr}
		\multirow{2}{*}{Method} && \multicolumn{4}{c}{Test error(\%)}  \\
		\cline{2-6}
		&$N_l$& $100$  & $600$ &  $1000$ & $3000$ \\
          	\hline\hline
            SVM~\cite[]{weston2012deep} &&23.44 & 8.85 & 7.77 & 4.21 \\
            TSVM~\cite[]{weston2012deep} &&16.81 & 6.16 & 5.38 & 3.45 \\
        	EmbedNN~\cite[]{weston2012deep} &&16.9& 5.97& 5.73 & 3.59 \\
			*MTC~\cite[]{rifai2011manifold}&&12.0&5.13& 3.64 & 2.57 \\
           PEA~\cite[]{bachman2014learning} &&10.79 & 2.44& 2.23& 1.91 \\
           *PEA~\cite[]{bachman2014learning}&&5.21 & 2.87&2.64& 2.30\\
           *DG~\cite[]{kingma2014semi}&&3.33 & 2.59& 2.40& 2.18\\
           *Ladder network~\cite[]{rasmus2015semi}&& 1.06& {}& 0.84&{} \\
           \hline
           Plain NN (MLE) &&21.98 & 9.16 & 7.25 & 4.32\\
           VAT~(ours)&&2.33 & 1.39& 1.36& 1.25\\
		\end{tabular}
    \end{subtable}
    }
    
    \scalebox{0.9}{
    \begin{subtable}[h]{1.0\textwidth}
        \centering
        \caption{SVHN \label{tab:SVHN} } 
		\begin{tabular}{lrr}
        	\multirow{2}{*}{Method} & \multicolumn{2}{c}{Test error(\%)} \\
            \cline{2-3}
		 &$N_l$&  $1000$ \\
          	\hline\hline
            TSVM~\cite[]{kingma2014semi} & &  66.55  \\
            *DG,M1+TSVM~\cite[]{kingma2014semi} & & 55.33 \\
            *DG,M1+M2~\cite[]{kingma2014semi} & & 36.02 \\
            \hline
            SVM (Gaussian kernel) && 63.28\\
            Plain NN (MLE) & & 43.21\\
            VAT~(ours) & & 24.63\\
		\end{tabular}
        \smallskip
    \end{subtable}
    }
    
    \scalebox{0.9}{
    \begin{subtable}[h]{1.0\textwidth}
        \centering
	    \caption{NORB  \label{tab:NORB}}
        \begin{tabular}{lrr}
        	\multirow{2}{*}{Method} & \multicolumn{2}{c}{Test error(\%)} \\
            \cline{2-3}
		 &$N_l$&  $1000$ \\
          	\hline\hline
            TSVM~\cite[]{kingma2014semi} & & 26.00  \\
            *DG,M1+TSVM~\cite[]{kingma2014semi} & & 18.79 \\
            \hline
            SVM (Gaussian kernel) && 23.62\\
            Plain NN (MLE)& & 20.00\\
            VAT~(ours) & & 9.88\\
		\end{tabular}
    \end{subtable}
    }
    \end{center}
\end{table}

\section{Discussion and Related Works}
Our VAT was motivated by the adversarial training \cite[]{goodfellow2014explaining}.
Adversarial training and VAT are similar in that they both use the local input-output relationship to smooth the model distribution in the corresponding neighborhood. In contrast, $L_2$ regularization does not use the local input-output relationship, and cannot introduce local smoothness to the model distribution. 
Increasing of the regularization constant in $L_2$ regularization can only intensify the global smoothing of the distribution, which results in higher training and generalization error. 
The adversarial training aims to train the model while keeping the average of the following value high:
\begin{eqnarray}
\min_{r} \{  \log p(y^{(n)}|x^{(n)}+r,\theta); \|r\|_p\leq\epsilon\}.
\end{eqnarray}
This makes the likelihood evaluated at the $n$-th labeled data point robust against $\epsilon-$perturbation applied to the input in its adversarial direction.   

PEA \cite[]{bachman2014learning}, on the other hand, used the model's sensitivity to random perturbation on input and hidden layers in their construction of the regularization function. PEA is similar to the random perturbation training of our experiment in that it aims to make the model distribution robust against random perturbation. Our VAT, however, outperforms PEA and random perturbation training. This fact is particularly indicative of the importance of the role of the hessian $H$ in VAT (Eq.(\ref{eq:second_taylor})). Because PEA and random perturbation training attempts to smooth the distribution into any arbitrary direction at every step of the update, it tends to make the variance of the loss function large in unnecessary dimensions. On the other hand, VAT always projects the perturbation in the principal direction of $H$, which is literally the principal direction into which the distribution is sensitive. 

Deep contractive network by Gu and Rigazio~\cite[]{gu2014towards} takes still another approach to smooth the model distribution.  Gu et al. introduced a penalty term based on the Frobenius norm of the Jacobian of the neural network's output $y = f(x, \theta)$ with respect to $x$.  
Instead of computing the computationally expensive full Jacobian, they approximated the Jabobian by the sum of the Frobenius norm of the Jacobian over every adjacent pairs of hidden layers.  The deep contractive network was, however, unable to significantly decrease the test error. 

Ladder network~\cite[]{rasmus2015semi} is a method that uses layer-wise denoising autoencoder. Their method is currently the best method in both supervised and semi-supervised learning for permutation invariant MNIST task.  
Ladder network seems to be  conducting a variation of manifold learning that extracts the knowledge of the local distribution of the inputs. \textcolor{black}{Still another classic way to include the information about the the input distribution is to use local smoothing of the data like Vicinal Risk Minimization (VRM)~\cite[]{chapelle2001vicinal}.} VAT, on the other hand, only uses the property of the conditional distribution $p(y | x, \theta)$, giving no consideration to the the generative process $p(x|\theta)$ nor to the full joint distribution $p(y, x| \theta).$
In this aspect, VAT can be complementary to the methods that explicitly model the input distribution. 
We might be able to improve VAT further by introducing the notion of manifold learning into its framework. 
\section{\textcolor{black}{Conclusion}}
\textcolor{black}{
Our experiments based on the synthetic datasets and the three real world dataset,  MNIST, SVHN and NORB indicate that the VAT is an effective method for both supervised and semi-supervised learning.  For the MNIST dataset, VAT outperformed all contemporary methods other than Ladder network, which is the current state of the art. VAT also outperformed current state of the art semi-supervised learning method for SVHN and NORB as well.  
We would also like to emphasize the simplicity of the method. With our approximation of LDS, VAT can be computed with relatively small computational cost. Also, models that relies heavily on generative models are dependent on many hyperparameters. VAT, on the other hand, has only two hyperparamaters, $\epsilon$ and $\lambda$. In fact,  our experiments worked sufficiently well with the optimization of one hyperparameter $\epsilon$ while fixing $\lambda=1$.   
}
\newpage

\bibliography{iclr2016_conference}

\begin{thebibliography}{28}
\providecommand{\natexlab}[1]{#1}
\providecommand{\url}[1]{\texttt{#1}}
\expandafter\ifx\csname urlstyle\endcsname\relax
  \providecommand{\doi}[1]{doi: #1}\else
  \providecommand{\doi}{doi: \begingroup \urlstyle{rm}\Url}\fi

\bibitem[Akaike(1998)]{akaike1998information}
Akaike, Hirotugu.
\newblock Information theory and an extension of the maximum likelihood
  principle.
\newblock In \emph{Selected Papers of Hirotugu Akaike}, pp.\  199--213.
  Springer, 1998.

\bibitem[Bachman et~al.(2014)Bachman, Alsharif, and
  Precup]{bachman2014learning}
Bachman, Phil, Alsharif, Ouais, and Precup, Doina.
\newblock Learning with pseudo-ensembles.
\newblock In \emph{Advances in Neural Information Processing Systems}, 2014.

\bibitem[Bastien et~al.(2012)Bastien, Lamblin, Pascanu, Bergstra, Goodfellow,
  Bergeron, Bouchard, and Bengio]{Bastien-Theano-2012}
Bastien, Fr{\'{e}}d{\'{e}}ric, Lamblin, Pascal, Pascanu, Razvan, Bergstra,
  James, Goodfellow, Ian~J., Bergeron, Arnaud, Bouchard, Nicolas, and Bengio,
  Yoshua.
\newblock Theano: new features and speed improvements.
\newblock Workshop on Deep Learning and Unsupervised Feature Learning at Neural
  Information Processing Systems, 2012.

\bibitem[Bengio(2009)]{bengio2009learning}
Bengio, Yoshua.
\newblock Learning deep architectures for ai.
\newblock \emph{Foundations and trends{\textregistered} in Machine Learning},
  2\penalty0 (1):\penalty0 1--127, 2009.

\bibitem[Bergstra et~al.(2010)Bergstra, Breuleux, Bastien, Lamblin, Pascanu,
  Desjardins, Turian, Warde-Farley, and Bengio]{bergstra+al:2010-scipy}
Bergstra, James, Breuleux, Olivier, Bastien, Fr{\'{e}}d{\'{e}}ric, Lamblin,
  Pascal, Pascanu, Razvan, Desjardins, Guillaume, Turian, Joseph, Warde-Farley,
  David, and Bengio, Yoshua.
\newblock Theano: a {CPU} and {GPU} math expression compiler.
\newblock In \emph{Proceedings of the Python for Scientific Computing
  Conference ({SciPy})}, June 2010.
\newblock Oral Presentation.

\bibitem[Chapelle et~al.(2001)Chapelle, Weston, Bottou, and
  Vapnik]{chapelle2001vicinal}
Chapelle, Olivier, Weston, Jason, Bottou, L{\'e}on, and Vapnik, Vladimir.
\newblock Vicinal risk minimization.
\newblock In \emph{Advances in Neural Information Processing Systems}, 2001.

\bibitem[Coates et~al.(2011)Coates, Ng, and Lee]{coates2011analysis}
Coates, Adam, Ng, Andrew~Y, and Lee, Honglak.
\newblock An analysis of single-layer networks in unsupervised feature
  learning.
\newblock In \emph{International Conference on Artificial Intelligence and
  Statistics}, pp.\  215--223, 2011.

\bibitem[Friedman et~al.(2001)Friedman, Hastie, and
  Tibshirani]{friedman2001elements}
Friedman, Jerome, Hastie, Trevor, and Tibshirani, Robert.
\newblock The elements of statistical learning.
\newblock \emph{Springer series in statistics Springer, Berlin}, 2001.

\bibitem[Glorot et~al.(2011)Glorot, Bordes, and Bengio]{glorot2011deep}
Glorot, Xavier, Bordes, Antoine, and Bengio, Yoshua.
\newblock Deep sparse rectifier neural networks.
\newblock In \emph{International Conference on Artificial Intelligence and
  Statistics}, pp.\  315--323, 2011.

\bibitem[Golub \& Van~der Vorst(2000)Golub and Van~der
  Vorst]{golub2000eigenvalue}
Golub, Gene~H and Van~der Vorst, Henk~A.
\newblock Eigenvalue computation in the 20th century.
\newblock \emph{Journal of Computational and Applied Mathematics}, 123\penalty0
  (1):\penalty0 35--65, 2000.

\bibitem[Goodfellow et~al.(2013)Goodfellow, Warde-Farley, Mirza, Courville, and
  Bengio]{goodfellow2013maxout}
Goodfellow, Ian~J, Warde-Farley, David, Mirza, Mehdi, Courville, Aaron, and
  Bengio, Yoshua.
\newblock Maxout networks.
\newblock In \emph{International Conference on Machine Learning}, 2013.

\bibitem[Goodfellow et~al.(2015)Goodfellow, Shlens, and
  Szegedy]{goodfellow2014explaining}
Goodfellow, Ian~J, Shlens, Jonathon, and Szegedy, Christian.
\newblock Explaining and harnessing adversarial examples.
\newblock In \emph{International Conference on Learning Representation}, 2015.

\bibitem[Gu \& Rigazio(2015)Gu and Rigazio]{gu2014towards}
Gu, Shixiang and Rigazio, Luca.
\newblock Towards deep neural network architectures robust to adversarial
  examples.
\newblock In \emph{International Conference on Learning Representation}, 2015.

\bibitem[Ioffe \& Szegedy(2015)Ioffe and Szegedy]{ioffe2015batch}
Ioffe, Sergey and Szegedy, Christian.
\newblock Batch normalization: Accelerating deep network training by reducing
  internal covariate shift.
\newblock In \emph{International Conference on Machine Learning}, 2015.

\bibitem[Jarrett et~al.(2009)Jarrett, Kavukcuoglu, Ranzato, and
  LeCun]{jarrett2009best}
Jarrett, Kevin, Kavukcuoglu, Koray, Ranzato, Marc'Aurelio, and LeCun, Yann.
\newblock What is the best multi-stage architecture for object recognition?
\newblock In \emph{Computer Vision, 2009 IEEE 12th International Conference
  on}, pp.\  2146--2153. IEEE, 2009.

\bibitem[Jia et~al.(2014)Jia, Shelhamer, Donahue, Karayev, Long, Girshick,
  Guadarrama, and Darrell]{jia2014caffe}
Jia, Yangqing, Shelhamer, Evan, Donahue, Jeff, Karayev, Sergey, Long, Jonathan,
  Girshick, Ross, Guadarrama, Sergio, and Darrell, Trevor.
\newblock Caffe: Convolutional architecture for fast feature embedding.
\newblock In \emph{Proceedings of the ACM International Conference on
  Multimedia}, pp.\  675--678. ACM, 2014.

\bibitem[Kingma \& Ba(2015)Kingma and Ba]{kingma2014adam}
Kingma, Diederik and Ba, Jimmy.
\newblock Adam: A method for stochastic optimization.
\newblock In \emph{International Conference on Learning Representation}, 2015.

\bibitem[Kingma et~al.(2014)Kingma, Mohamed, Rezende, and
  Welling]{kingma2014semi}
Kingma, Diederik, Mohamed, Shakir, Rezende, Danilo~Jimenez, and Welling, Max.
\newblock Semi-supervised learning with deep generative models.
\newblock In \emph{Advances in Neural Information Processing Systems}, 2014.

\bibitem[LeCun et~al.(2015)LeCun, Bengio, and Hinton]{lecun2015deep}
LeCun, Yann, Bengio, Yoshua, and Hinton, Geoffrey.
\newblock Deep learning.
\newblock \emph{Nature}, 521\penalty0 (7553):\penalty0 436--444, 2015.

\bibitem[Nair \& Hinton(2010)Nair and Hinton]{nair2010rectified}
Nair, Vinod and Hinton, Geoffrey~E.
\newblock Rectified linear units improve restricted boltzmann machines.
\newblock In \emph{International Conference on Machine Learning}, 2010.

\bibitem[Pearlmutter(1994)]{pearlmutter1994fast}
Pearlmutter, Barak~A.
\newblock Fast exact multiplication by the hessian.
\newblock \emph{Neural computation}, 6\penalty0 (1):\penalty0 147--160, 1994.

\bibitem[Rasmus et~al.(2015)Rasmus, Valpola, Honkala, Berglund, and
  Raiko]{rasmus2015semi}
Rasmus, Antti, Valpola, Harri, Honkala, Mikko, Berglund, Mathias, and Raiko,
  Tapani.
\newblock Semi-supervised learning with ladder network.
\newblock \emph{arXiv preprint arXiv:1507.02672}, 2015.

\bibitem[Rifai et~al.(2011)Rifai, Dauphin, Vincent, Bengio, and
  Muller]{rifai2011manifold}
Rifai, Salah, Dauphin, Yann~N, Vincent, Pascal, Bengio, Yoshua, and Muller,
  Xavier.
\newblock The manifold tangent classifier.
\newblock In \emph{Advances in Neural Information Processing Systems}, 2011.

\bibitem[Srivastava et~al.(2014)Srivastava, Hinton, Krizhevsky, Sutskever, and
  Salakhutdinov]{srivastava2014dropout}
Srivastava, Nitish, Hinton, Geoffrey, Krizhevsky, Alex, Sutskever, Ilya, and
  Salakhutdinov, Ruslan.
\newblock {Dropout: A simple way to prevent neural networks from overfitting}.
\newblock \emph{The Journal of Machine Learning Research}, 15\penalty0
  (1):\penalty0 1929--1958, 2014.

\bibitem[Tokui et~al.(2015)Tokui, Oono, Hido, and Clayton]{tokuichainer}
Tokui, Seiya, Oono, Kenta, Hido, Shohei, and Clayton, Justin.
\newblock Chainer: a next-generation open source framework for deep learning.
\newblock In \emph{Workshop on Machine Learning Systems at Neural Information
  Processing Systems}, 2015.

\bibitem[Wahba(1990)]{wahba1990spline}
Wahba, Grace.
\newblock Spline models for observational data.
\newblock \emph{Siam}, 1990.

\bibitem[Watanabe(2009)]{watanabe2009algebraic}
Watanabe, Sumio.
\newblock Algebraic geometry and statistical learning theory.
\newblock \emph{Cambridge University Press}, 2009.

\bibitem[Weston et~al.(2012)Weston, Ratle, Mobahi, and
  Collobert]{weston2012deep}
Weston, Jason, Ratle, Fr{\'e}d{\'e}ric, Mobahi, Hossein, and Collobert, Ronan.
\newblock Deep learning via semi-supervised embedding.
\newblock In \emph{Neural Networks: Tricks of the Trade}, pp.\  639--655.
  Springer, 2012.

\end{thebibliography}
\bibliographystyle{iclr2016_conference}

\appendix

\section{Appendix:Details of experimental settings}
\subsection{Supervised binary classification for synthetic datasets}\label{apd:syn_exp_set}
We provide more details of experimental settings on synthetic datasets. 
We list the search space for the hyperparameters below:
\vspace{-3mm}
\begin{tight_itemize}
\item $L_2$ regularization: regularization coefficient $\lambda=\{1e-4,\cdots, 200\}$
\item Dropout (only for the input layer): dropout rate $p(z) = \{0.05,\cdots,0.95\}$
\item Random perturbation training: $\epsilon=\{0.2,\cdots,4.0\}$
\item Adversarial training (with $L_{\infty}$ norm constraint): $\epsilon=\{0.01,\cdots,0.2\}$
\item Adversarial training (with $L_{2}$ norm constraint):  $\epsilon=\{0.1,\cdots,2.0\}$
\item VAT: $\epsilon=\{0.1,\cdots,2.0\}$
\end{tight_itemize}
\vspace{-3mm}
All experiments with random perturbation training, adversarial training and VAT were conducted with $\lambda = 1$.
As for the training we used stochastic gradient descent (SGD) with a moment method. 
When $J(\theta)$ is the objective function,
the moment method augments the simple update in the SGD with a term dependent on the previous update $\Delta \theta_{i-1}$:
\begin{eqnarray}
\Delta\theta_{i} = \mu_i\Delta\theta_{i-1} + ( 1-\mu_i)\gamma_i \frac{\partial}{\partial \theta} J(\theta). \label{eq:theta}
\end{eqnarray}
In the expression above, $\mu_i \in [0,1)$ stands for the strength of the momentum, and $\gamma_i$ stands for the learning rate. 
In our experiment, we used $\mu_i= 0.9$, and exponentially decreasing $\gamma_i$ with rate $0.995$.  As for the choice of
$\gamma_1$, we used $1.0$.  
We trained the NNs with 1,000 parameter updates.

\subsection{Supervised classification for the MNIST dataset}\label{apd:mnist_exp_set}
We provide more details of experimental settings on supervised classification for the MNIST dataset.
Following lists summarizes the ranges from which we searched for the best hyperparameters of each regularization method:
\vspace{-3mm}
\begin{tight_itemize}
\item Random perturbation training: $\epsilon = \{5.0,\cdots,15.0\}$
\item Adversarial training (with $L_{\infty}$ norm constraint): $\epsilon=\{0.05,\cdots,0.1\}$
\item Adversarial training (with $L_{2}$ norm constraint): $\epsilon=\{1.0,\cdots,3.0\}$
\item VAT: $\epsilon =\{1.0,\cdots,3.0\}$, $I_p=1$
\end{tight_itemize}
\vspace{-3mm}
All experiments were conducted with $\lambda = 1$.
The training was conducted by mini-batch SGD based on ADAM~\cite[]{kingma2014adam}.
We chose the mini-batch size of 100, and used the default values of \cite{kingma2014adam} for the tunable parameters
of ADAM. We trained the NNs with 50,000 parameter updates. As for the base learning rate in validation, we selected the initial value of $0.002$ and adopted the schedule of exponential decay with rate $0.9$ per 500 updates. 
After the hyperparameter determination, we trained the NNs over 60,000 parameter updates.
For the learning coefficient, we used the initial value of $0.002$ and adopted the schedule of exponential decay with rate $0.9$ per 600 updates.

\subsection{Semi-supervised classification for the MNIST dataset}\label{apd:semi-mnist_exp_set}
We provide more details of experimental settings on semi-supervised classification for the MNIST dataset.
We searched for the best hyperparameter $\epsilon$ from  $\{0.2,0.3,0.4\}$ in the $N_l = 100$ case. Best $\epsilon$ was selected from $\{1.5,2.0,2.5\}$ for all other cases.  All experiments were conducted with $\lambda = 1$ and $I_p =1$. For the optimization method, we again used ADAM-based minibatch SGD with the same hyperparameter values as those in the supervised setting. We note that, in the computation of ADAM, the likelihood term can be computed from labeled data only.
We therefore used two separate minibatches at each step: one minibatch of size 100 from labeled samples for the computation of the likelihood term, and another minibatch of size 250 from both labeled and unlabeled samples for computing the regularization term. We trained the NNs over 50,000 parameter updates. For the learning rate, we used the initial value of $0.002$ and adopted the schedule of exponential decay with rate $0.9$ per 500 updates.

\subsection{Semi-supervised classification for the SVHN and NORB dataset}\label{apd:semi-svhn_norb_exp_set}
{\color{black}
We provide the details of the numerical experiment we conducted for the SVHN and NORB dataset.\\
The SVHN dataset consists of $32\times 32 \times 3$ pixel RGB images of housing numbers and their corresponding labels (0-9), and  the number of  training samples and test samples within the dataset are 73,257 and 26,032, 
respectively. To simplify the experiment, we down sampled the images from  $32\times 32 \times 3$ to 
$16 \times 16 \times 3$. We vectorized each image to 768 dimensional vector, and applied whitening~\cite[]{coates2011analysis} to the dataset.  We reserved 1000 dataset for validation. From the rest, we used 1000 dataset as labeled dataset in semi-supervised training. We repeated the experiment 10 times with different choice of labeled dataset and validation dataset.\\
The NORB dataset consists of $2 \times  96\times 96$ pixel gray images of 50 different objects 
and their corresponding labels (cars, trucks, planes, animals, humans).   The number of training samples and test samples constituting the dataset are 24,300. We downsampled the images from  $2 \times  96\times 96$ to $2 \times 32\times 32$.  We vectorized each image to 2048 dimensional vector and applied whitening. We reserved 1000 dataset for validation. From the rest, we used 1000 dataset as labeled dataset in semi-supervised training. We repeated the experiment 10 times with different choice of labeled dataset and validation dataset. 

For both SVHN and NORB, we used neural network with the number of hidden nodes given by $(1200,600,300,150,150)$. In our setup, these two dataset preferred deeper network than the network we used for the MNIST dataset. We used ReLU for the activation function. As for the hyperparameters $\epsilon$, we conducted grid search over the range $\{1.0,1.5,\cdots,4.5,5.0\}$, and we used $\lambda = 1$. For power iteration, we used $I_p = 1$.  

We used ADAM based minibach SGD with the same hyperparameter values as the MNIST. We chose minibatch size of 100. Thus one epoch for SVHN completes with $\lceil 73,257 / 100 \rceil = 753$ rounds of minibatches.  
For the learning rate, we used the initial value of $0.002$ with exponential decay of rate $0.9$ per epoch. 
We trained NNs with 100 epochs.}
 \end{document}